\documentclass[pdflatex,sn-basic]{sn-jnl}

\usepackage{graphicx}%
\usepackage{multirow}%
\usepackage{amsmath,amssymb,amsfonts}%
\usepackage{amsthm}%
\usepackage{mathrsfs}%
\usepackage[title]{appendix}%
\usepackage{xcolor}%
\usepackage{textcomp}%
\usepackage{manyfoot}%
\usepackage{booktabs}%
\usepackage{algorithm}%
\usepackage{algorithmicx}%
\usepackage{algpseudocode}%
\usepackage{listings}%

\theoremstyle{thmstyleone}%
\theoremstyle{thmstyletwo}%

\theoremstyle{thmstylethree}%

\begin{document}

\title[Article Title]{UniDCF: A Foundation Model for Comprehensive Dentocraniofacial Hard Tissue Reconstruction}


\author[1]{\fnm{Chunxia} \sur{Ren}}\email{renchunxia@bupt.edu.cn}
\equalcont{These authors contributed equally to this work.}

\author[2]{\fnm{Ning} \sur{Zhu}}\email{dentzn@163.com}
\equalcont{These authors contributed equally to this work.}

\author[4]{\fnm{Yue} \sur{Lai}}\email{ly2882881@163.com}
\equalcont{These authors contributed equally to this work.}

\author[4]{\fnm{Gui} \sur{Chen}}\email{chengui723@163.com}
\author[3]{\fnm{Ruijie} \sur{Wang}}\email{pkuwangruijie@126.com}
\author[2]{\fnm{Yangyi} \sur{Hu}}\email{huyangyi10@gmail.com}
\author[5]{\fnm{Suyao} \sur{Liu}}\email{2022210074@bupt.cn}
\author[5]{\fnm{Shuwen} \sur{Mao}}\email{2022210049@bupt.cn}

\author*[3]{\fnm{Hong} \sur{Su}}\email{suhong2004@163.com}
\author*[2]{\fnm{Yu} \sur{Zhang}}\email{zhang76yu@163.com}
\author*[1]{\fnm{Li} \sur{Xiao}}\email{andrew.lxiao@gmail.com}

\affil*[1]{\orgdiv{School of Artificial Intelligence}, \orgname{Beijing University of Posts and Telecommunications}, \orgaddress{\street{Xitucheng Road 10, Haidian District}, \city{Beijing}, \postcode{100876}, \country{P.R. China}}}

\affil*[2]{\orgdiv{Department of Oral Implantology}, \orgname{Peking University School and Hospital of Stomatology \& National Center for Stomatology \& National Clinical Research Center for Oral Diseases \& National Engineering Research Center of Oral Biomaterials and Digital Medical Devices}, \orgaddress{\street{No.22, Zhongguancun South Avenue, Haidian District}, \city{Beijing}, \postcode{100081}, \country{P.R. China}}}

\affil*[3]{\orgdiv{First Clinical Division}, \orgname{Peking University School and Hospital of Stomatology \& National Clinical Research Center for Oral Disease \& National Engineering Research Center of Oral Biomaterials and Digital Medical Devices \& Beijing Key Laboratory of Digital Stomatology}, \orgaddress{\street{No.37A, Xishiku Avenue, Xicheng District}, \city{Beijing}, \postcode{100034}, \country{P.R. China}}}

\affil[4]{\orgdiv{Department of Orthodontics}, \orgname{Cranial-Facial Growth and Development Center \& Peking University School and Hospital of Stomatology \& National Center of Stomatology \& National Clinical Research Center for Oral Diseases \& National Engineering Laboratory for Digital and Material Technology of Stomatology \& Beijing Key Laboratory for Digital Stomatology \& Research Center of Engineering and Technology for Computerized Dentistry Ministry of Health \& NMPA Key Laboratory for Dental Materials}, \orgaddress{\street{No.22, Zhongguancun South Avenue, Haidian District}, \city{Beijing}, \postcode{100081}, \country{P.R. China}}}

\affil[5]{\orgdiv{School of Future}, \orgname{Beijing University of Posts and Telecommunications}, \orgaddress{\street{Xitucheng Road 10, Haidian District}, \city{Beijing}, \postcode{100876}, \country{P.R. China}}}


\abstract{Dentocraniofacial hard tissue defects profoundly affect patients’ physiological functions, facial aesthetics, and psychological well-being, posing significant challenges for precise reconstruction. Current deep learning models are limited to single-tissue scenarios and modality-specific imaging inputs, resulting in poor generalizability and trade-offs between anatomical fidelity, computational efficiency, and cross-tissue adaptability. Here we introduce UniDCF, a unified framework capable of reconstructing multiple dentocraniofacial hard tissues through multimodal fusion encoding of point clouds and multi-view images. By leveraging the complementary strengths of each modality and incorporating a score-based denoising module to refine surface smoothness, UniDCF overcomes the limitations of prior single-modality approaches. We curated the largest multimodal dataset, comprising intraoral scans, CBCT, and CT from 6,609 patients, resulting in 54,555 annotated instances. Evaluations demonstrate that UniDCF outperforms existing state-of-the-art methods in terms of geometric precision, structural completeness, and spatial accuracy. Clinical simulations indicate UniDCF reduces reconstruction design time by 99\% and achieves clinician-rated acceptability exceeding 94\%. Overall, UniDCF enables rapid, automated, and high-fidelity reconstruction, supporting personalized and precise restorative treatments, streamlining clinical workflows, and enhancing patient outcomes.}




\maketitle

\section{Introduction}\label{sec1}
Dentocraniofacial hard tissues—including the skull, teeth, and jaws—form the structural basis of facial morphology and support essential functions such as mastication, speech, and neurological protection \cite{koons2020materials}. Millions of individuals worldwide suffer annually from defects in these tissues because of trauma, tumor resection, congenital malformations, or diseases such as severe caries and periodontitis \cite{moncal2021intra, li20233d, chen2025burden, ccirak2025craniofacial}. These defects compromise both physiological function and facial aesthetics, leading to psychological distress, impaired social interactions, and diminished quality of life \cite{aghali2021craniofacial, tooth2021, tooth2023}. As the prevalence and complexity of these conditions grow, there is an urgent clinical need for precise, efficient, and personalized reconstruction solutions.

\begin{figure}[htbp]
\centering
\includegraphics[width=0.93\textwidth]{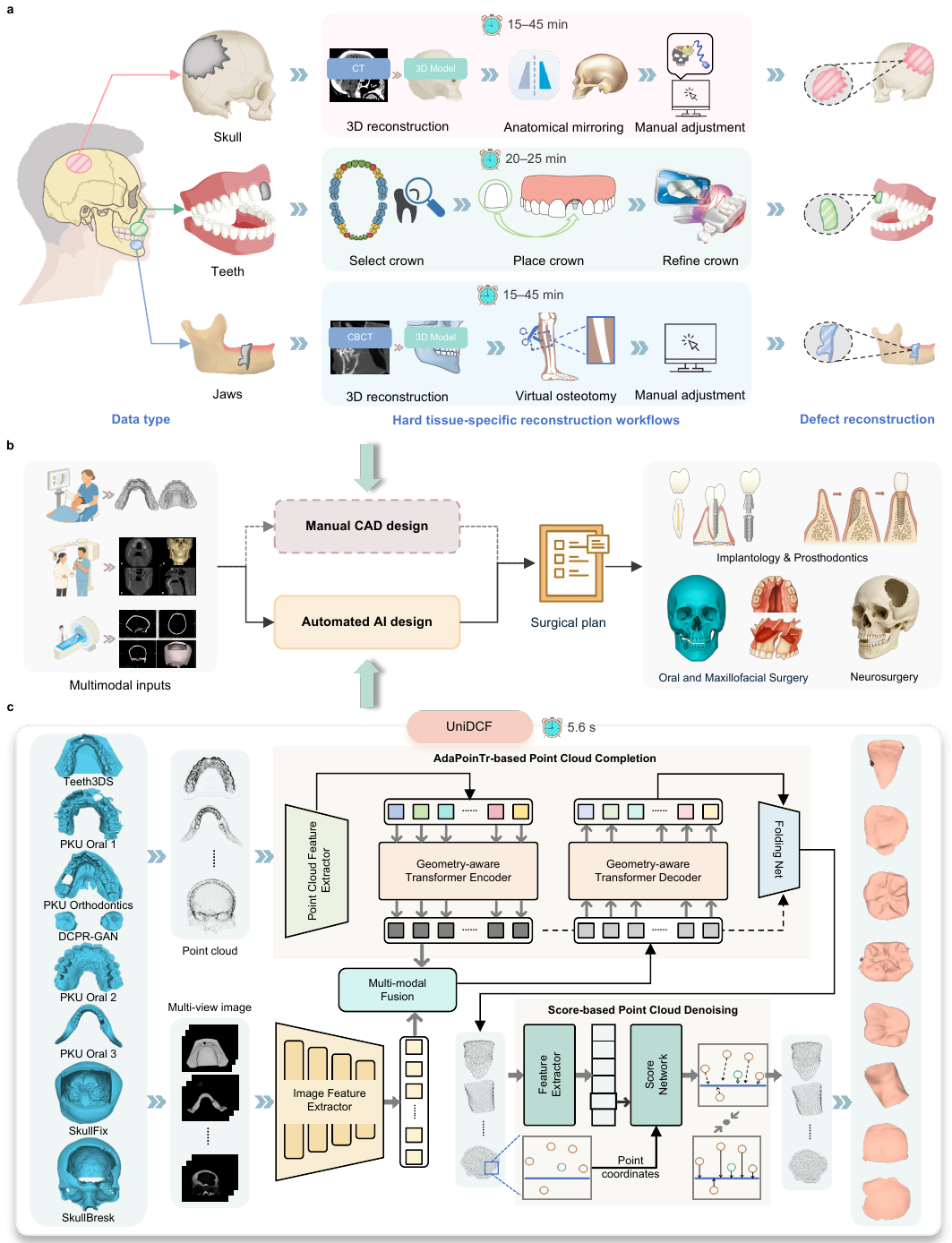}
\caption{Overview of the dentocraniofacial hard tissue reconstruction workflow and UniDCF architecture.
a, Conventional CAD-based clinical workflow for dentocraniofacial reconstruction. The processes for skull, teeth, and jaws reconstructions are shown separately, with the associated manual design time indicated. b, A unified clinical scenarios for dentocraniofacial reconstruction, encompassing multimodal image acquisition, treatment planning using CAD or AI-assisted approaches, and downstream surgical interventions including implant prosthodontic procedures, such as crown design and bone augmentation, oral and maxillofacial surgery (e.g., cleft lip and palate repair), and neurosurgery. This panel illustrates the integrative potential and clinical applicability of a unified reconstruction framework.
c, The architecture of UniDCF. Built on the AdaPoinTr point cloud completion backbone, UniDCF takes sparse point clouds and multi-view grayscale images as input and employs cross-modal fusion and score-based denoising to generate anatomically accurate dentocraniofacial reconstructions.
}\label{fig1}
\end{figure}

Advances in medical imaging, computer-aided design (CAD), and three-dimensional (3D) printing \cite{CAD2016,cad2018,cad2023, copelli2024bone,abdulkarim2024impact} have shifted dentocraniofacial reconstruction from manual, experience-dependent workflows to digital, precision-based paradigms. Conventional CAD methods \cite{mehl2005new,zhang2015reconstruction,zhang2017computer} typically rely on selecting generic templates—such as dental crown libraries or normative cranial models—and adapting them to patient-specific defects via mirroring or template matching, followed by manual refinements (Fig. \ref{fig1}a). While more accurate than fully manual techniques, these approaches remain labor intensive, depend heavily on clinical expertise, and often fall short in delivering truly individualized outcomes. Additionally, separate workflows and imaging modalities for m multiple dentocraniofacial hard tissues exacerbate fragmentation, impeding clinical integration and consistency.

Artificial intelligence (AI) offers a promising avenue to overcome these limitations \cite{ding2023artificial, aljulayfi2024potential, wu2025comparison, juneja2025comprehensive}. AI models can learn diagnostic and restorative features, and have shown success in automating tasks such as crown prosthodontics and image analysis \cite{lee2018diagnosis, krois2021generalizability, cui2022fully, tassoker2024performance}. However, most existing methods remain narrowly focused, limiting their adaptability to the full spectrum of dentocraniofacial reconstruction needs. Consequently, there remains a pressing need for a unified foundational AI framework capable of reconstructing multiple dentocraniofacial hard tissues concurrently.

Achieving comprehensive dentocraniofacial hard tissue reconstruction faces two fundamental challenges. First, the development of generalizable AI models is hindered by the lack of large-scale, multimodal, and multi-tissue datasets, which are essential for learning transferable anatomical representations across diverse dentocraniofacial structures. As a result, most current methods typically target single-tissue domains (e.g., teeth or skull) and train on modality-specific data (e.g., intraoral scans, cone-beam computed tomography (CBCT), or CT), severely limiting their applicability across heterogeneous anatomical regions. Second, current single-modality methods struggle to balance anatomical fidelity, computational efficiency, and cross-tissue adaptability. Specifically, pixel-based 2D models \cite{tian2022dual, DCPR-GAN} inadequately capture spatially coherent 3D morphology; voxel-based 3D models \cite{kwarciak2023deep, wodzinski2024automatic} are computationally intensive and difficult to scale for high-resolution reconstruction tasks; and although point cloud–based methods \cite{hosseinimanesh2023mesh, point2} are computationally efficient, their outputs are susceptible to local noise, resulting in limited robustness and poor surface smoothness.

To overcome these challenges, we propose UniDCF, a foundation model for comprehensive dentocraniofacial hard tissue reconstruction. Designed to operate across tissues and imaging modalities, UniDCF addresses both the lack of interoperable data formats and the limitations of single-modality inference. To standardize heterogeneous inputs, our preprocessing pipeline (Fig. \ref {fig1}b, c) converts intraoral scans, CBCT, and CT data into two unified representations: sparse point clouds and multi-view grayscale images. Sparse point clouds retain global spatial relationships and anatomical topology, while multi-view images encode local curvature and morphological features. UniDCF integrates these complementary modalities through a multimodal fusion architecture built on the AdaPoinTr \cite{AdaPoinTr} backbone, which combines transformer-based geometric reasoning with cross-modal feature encoding. Additionally, UniDCF incorporates a score-based denoising module \cite{luo2021score} to further reduce surface irregularities and enhance geometric smoothness. This unified architecture enables end-to-end reconstruction across clinical scenarios from dental crowns to craniofacial bone implants. To support generalizable learning, we curated the largest integrated dentocraniofacial hard tissue dataset to date, comprising 54,555 samples from 6,609 patients. This dataset integrates multiple hard tissue types—including skull, teeth, and jaws—spanning diverse anatomical regions, clinical departments (e.g., implantology, orthodontics, neurosurgery), and imaging modalities. UniDCF improves both reconstruction precision and workflow automation by unifying architectural design and scaling data integration, laying the foundation for clinically scalable and anatomically robust AI-driven solutions in dentocraniofacial medicine.

We validated UniDCF through extensive experimental and clinical evaluations on the large-scale, multimodal dataset. Comparative analysis across datasets of varying size, modality, and source underscores the critical role of data diversity in model generalizability. Benchmarks against state-of-the-art methods show that UniDCF achieves superior geometric accuracy, structural completeness, and cross-tissue adaptability. Clinically, UniDCF reduces manual design time by over 99\%, with 94.2\% of reconstructions deemed clinically acceptable by expert evaluators based on anatomical fidelity to natural structures. Together, these findings highlight UniDCF's potential to transform dentocraniofacial reconstruction by enhancing precision, scalability, and clinical efficiency. The code and dataset will be made publicly available upon publication to support transparency, reproducibility, and further research in multimodal dentocraniofacial reconstruction.

\section{Results}\label{sec2}
\subsection{Dataset characteristics}\label{subsec21}
\begin{figure}[htbp]
\centering
\includegraphics[width=0.99\textwidth]{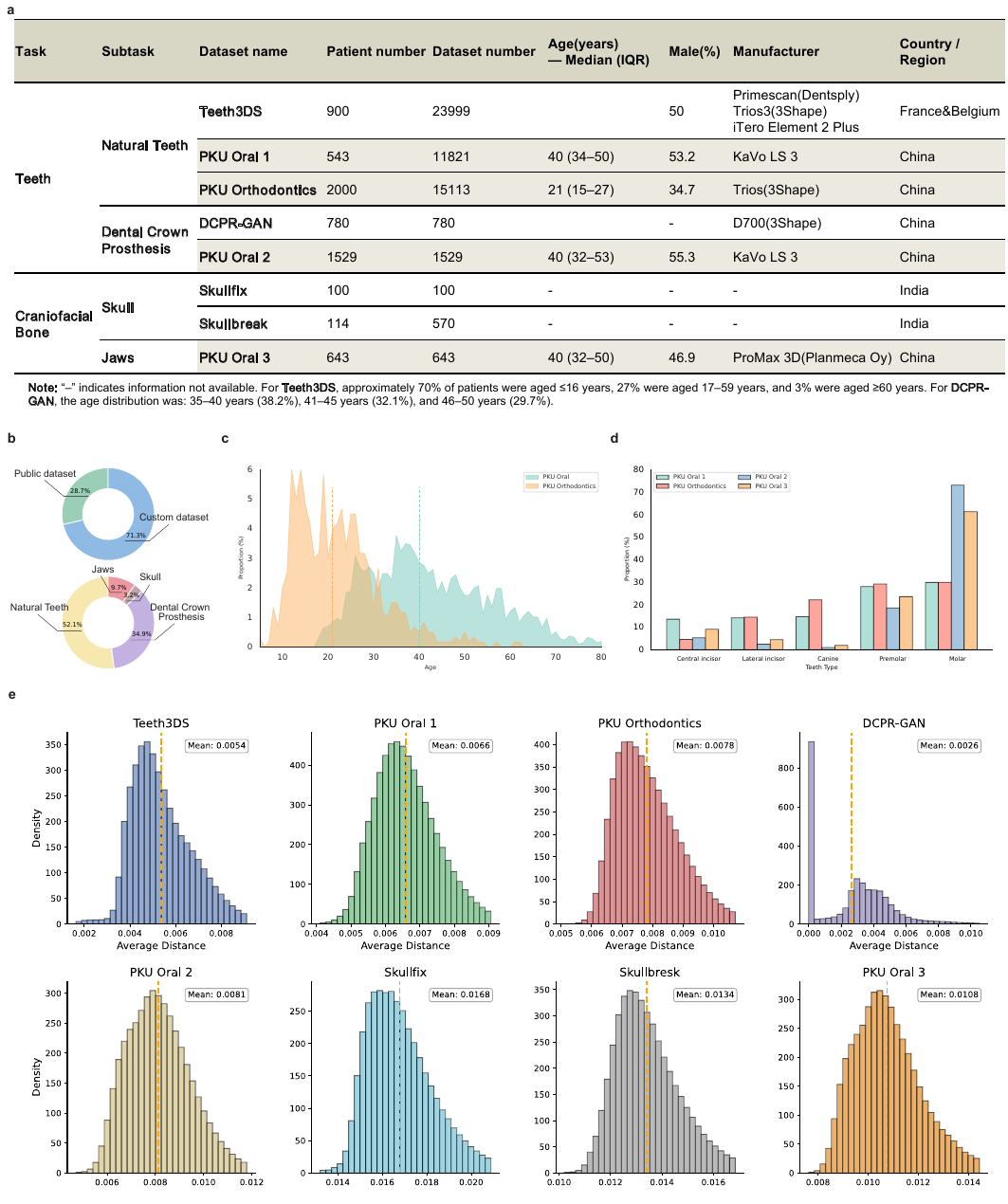}
\caption{
Overview and diversity analysis of the integrated dentocraniofacial hard tissue dataset.
a, Summary table describing dataset characteristics, including patient demographics, imaging modalities, scanner types, and geographic distribution.
b, Donut charts showing the number of patients from custom-collected (Custom dataset) and publicly available (Public dataset) sources (top), and the distribution across anatomical categories (bottom): natural teeth, dental crown prosthesis, skull, and jaws.
c, Stacked area chart comparing age distribution differences between orthodontic and implant patients from the Peking University School and Hospital of Stomatology, highlighting age concentration between 10–30 years for orthodontics and 30–50 years for implants.
d, Grouped bar chart showing the distribution of hard tissue defects across dental arches in four custom datasets. PKU Oral 1 exhibits a balanced distribution, while PKU Oral 2 and PKU Oral 3 are skewed toward molar region defects.
e, Histogram of point cloud sparsity across datasets. Craniofacial bone datasets have higher average point density than teeth datasets, with DCPR-GAN exhibiting the lowest point count.
}\label{fig2}
\end{figure}

We assembled a comprehensive dataset comprising 54,555 samples from 6,609 patients, covering a broad spectrum of dentocraniofacial hard tissues, including teeth, jaws, and skull. This large-scale and clinically diverse resource enables anatomically precise reconstructions tailored to the needs of various specialties, such as orthodontics, prosthodontics, implantology, and cranio-maxillofacial surgery. The dataset integrates four public datasets (Teeth3DS \cite{miccai}, DCPR-GAN \cite{DCPR-GAN}, Skullfix, and Skullbreak \cite{skull}) with four retrospective clinical cohorts from the Department of Oral Implantology and Orthodontics at the eking University School and Hospital of Stomatology (PKU Oral 1, PKU Orthodontics, PKU Oral 2, and PKU Oral 3), the largest tertiary oral healthcare center in China. To our knowledge, this constitutes the most diverse and clinically representative dataset in the field (Fig. \ref {fig2}a-d). 

Spanning a diverse array of imaging modalities, tissue types, and clinical scenarios, the dataset provides robust support for dentocraniofacial reconstruction: Natural teeth datasets (Teeth3DS, PKU Oral 1, PKU Orthodontics) consist of full-arch intraoral scans from patients across a wide age range and varying degrees of dental crowding. These data facilitate orthodontic planning, aesthetic analysis, and implant site evaluation by enabling accurate reconstruction of missing or morphologically compromised dental crowns. Dental crown prosthesis datasets (DCPR-GAN, PKU Oral 2) include prosthetic crowns across incisor, canine, premolar and molar, supporting intelligent denture design with optimized functional–aesthetic integration, biomechanical performance, and long-term stability. Skull datasets (Skullfix, Skullbreak) include diverse skull defects in terms of size, morphology, and location. This data is essential for restoring skull structure, protecting neurological function, ensuring facial symmetry, and alleviating psychological burden in patients with craniofacial deformities. Jaws dataset (PKU Oral 3) encompasses a wide range of bone resorption patterns, offering crucial anatomical information for implant planning, bone augmentation, and improved surgical accuracy and safety. To ensure compatibility with the UniDCF framework, all datasets underwent standardized preprocessing. This included conversion of raw multimodal inputs into sparse point clouds and corresponding multi-view grayscale projections (XYZ views). The distribution of point densities across datasets is shown in Fig. \ref {fig2}e.

\subsection{Experimental Results}\label{subsec22}
\subsubsection{Evaluating Cross-task Generalization Driven by Multi-tissue Dataset Diversity}\label{subsubsec221}
\begin{figure}[htbp]
\centering
\includegraphics[width=\textwidth]{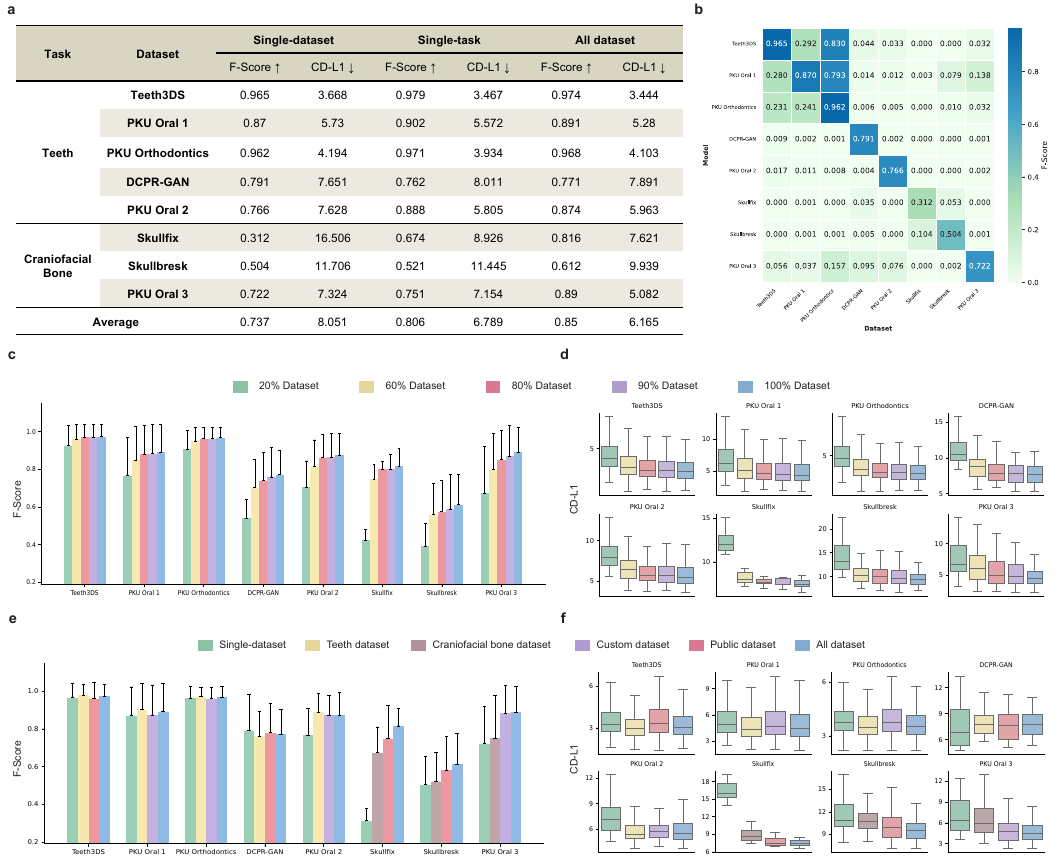}
\caption{
Model performance under diverse data configurations.
a, Quantitative comparison table of model performance trained under three data configurations: Single-dataset, Single-task, and All dataset. Results are reported as F1-Score and CD-L1 across evaluation datasets.
b, Cross-dataset generalization of independently trained models (Single-dataset), shown as a confusion matrix based on F1-Score.
c, Grouped bar chart showing F1-Score for different data volume ratios (20\%, 60\%, 80\%, 90\%, 100\%) of All dataset.
d, Box plots showing CD-L1 values under the same volume scaling conditions as in c.
e–f, Performance comparisons (F1-Score and CD-L1) under varied dataset mixing strategies: Single-dataset, Single-task (e.g., teeth dataset vs. Craniofacial bone dataset), Combined-task (grouped by source: Public dataset vs. Custom dataset), and All dataset conditions.}\label{fig3}
\end{figure}
To systematically assess the impact of dataset diversity on cross-task generalization, we reformulated teeth and craniofacial bone reconstruction as two structurally distinct yet semantically related tasks under the broader category of hard tissue repair. This task-level reformulation enabled a comprehensive investigation into how anatomical diversity—spanning data source, scale, and tissue domain—affects feature sharing and knowledge transfer across tasks. We quantified geometric accuracy using Chamfer Distance (CD-L1) and point-level reconstruction quality using the F1-Score. CD-L1 was computed as the L1-norm distance between predicted and ground-truth point clouds, multiplied by 1,000 for readability.
We examined model generalization under three data integration strategies: (1) independent training on individual datasets (Single-dataset), (2) joint training within task-specific groups (Single-task, involving either teeth-only or craniofacial bone-only reconstruction tasks), and (3) unified training across all datasets (All dataset). All models were evaluated using a consistent test set for comparability.

As shown in Fig. \ref{fig3}a, the model trained on the All dataset achieved the highest overall performance, with an average F1-Score of 0.85 and an average CD-L1 of 6.165—substantially outperforming the Single-task and Single-dataset variants. These results highlight the benefits of multi-tissue learning and anatomical diversity in improving generalizability. The Single-task model also outperformed the Single-dataset model, with an average F1-Score gain of 0.069 and an average CD-L1 reduction of 1.262, suggesting that task-level data aggregation enhances intra-task representation learning.
Dataset-specific analysis further confirmed this trend: models trained on the All dataset and Single-task configurations consistently surpassed the Single-dataset models for most datasets, reflecting the advantages of broader learning contexts. Notably, the All dataset model achieved significant improvements in craniofacial bone tasks (F1-Score increases $\geqslant$ 0.091; CD-L1 reductions $\geqslant$ 1.305), because of structural priors learned from the larger teeth datasets. In contrast, performance in teeth tasks—already well-supported by abundant training data—showed similar performance, potentially due to interference from morphologically dissimilar craniofacial bone inputs. These asymmetric trends suggest that multi-task integration is particularly advantageous for clinically important but data-scarce tasks, where structural transfer promotes more robust representations.
The DCPR-GAN dataset showed an exception, where the Single-dataset model performed slightly better than Single-task and All dataset. The atypical data distribution of the DCPR-GAN dataset likely caused this deviation. Unlike other teeth datasets with complete arches, DCPR-GAN consists of partial segments centered on adjacent teeth flanking missing crowns, limiting the benefits of task- or dataset-level knowledge transfer.

We next evaluated cross-dataset generalization by independently training models on one dataset and testing them across the others (Fig. \ref {fig3}b). As expected, performance declined because of domain shifts in anatomical geometry and acquisition standards. Natural teeth datasets (Teeth3DS, PKU Oral 1, PKU Orthodontics) exhibited moderate generalizability, likely due to shared anatomical priors. For instance, models trained on Teeth3DS and PKU Oral 1 generalized well to PKU Orthodontics, with F1-Scores of 0.83 and 0.793, respectively. Conversely, models trained on PKU Orthodontics failed to generalize effectively, due to low intra-dataset variability and feature space overfitting. Similarly, models trained on dental crown prosthesis (DCPR-GAN, PKU Oral 2) or craniofacial
bone datasets demonstrated poor cross-dataset generalization—even within their respective tasks—highlighting the limitations imposed by small, heterogeneous training sets. To assess the influence of dataset scale, we incrementally trained models using 20\%, 60\%, 80\%, 90\%, and 100\% of the full dataset (Fig. \ref {fig3}c, d). Performance improved substantially between 20\% and 60\% training data, particularly in craniofacial bone tasks. However, performance gains plateaued beyond the 80\% threshold, indicating diminishing returns with increasing dataset size. These findings suggest that model generalizability is highly sensitive to dataset scale in early training phases, but saturates as data volume increases.

Finally, we compared the effects of different dataset composition strategies on model performance (Fig. \ref{fig3}e, f). Specifically, we evaluated models trained on individual datasets (Single-dataset), task-level groups (teeth dataset vs. Craniofacial bone dataset), data source groups (Public dataset vs. Custom datasets), and the entire dataset (All dataset). For teeth tasks, high-precision reconstruction was achieved using large-scale teeth datasets alone, and the inclusion of craniofacial bone data resulted in negligible performance changes. In contrast, craniofacial bone reconstruction—limited by smaller sample size—benefited significantly from teeth dataset inclusion, likely due to shared morphological priors. These results underscore the clinical value of cross-domain feature transfer and highlight the importance of multi-tissue dataset integration in enabling robust, generalizable models for comprehensive dentocraniofacial hard tissue reconstruction.

\subsubsection{Comparative and Visual Analysis}\label{subsubsec222}
\begin{figure}[htbp]
\centering
\includegraphics[width=0.95\textwidth]{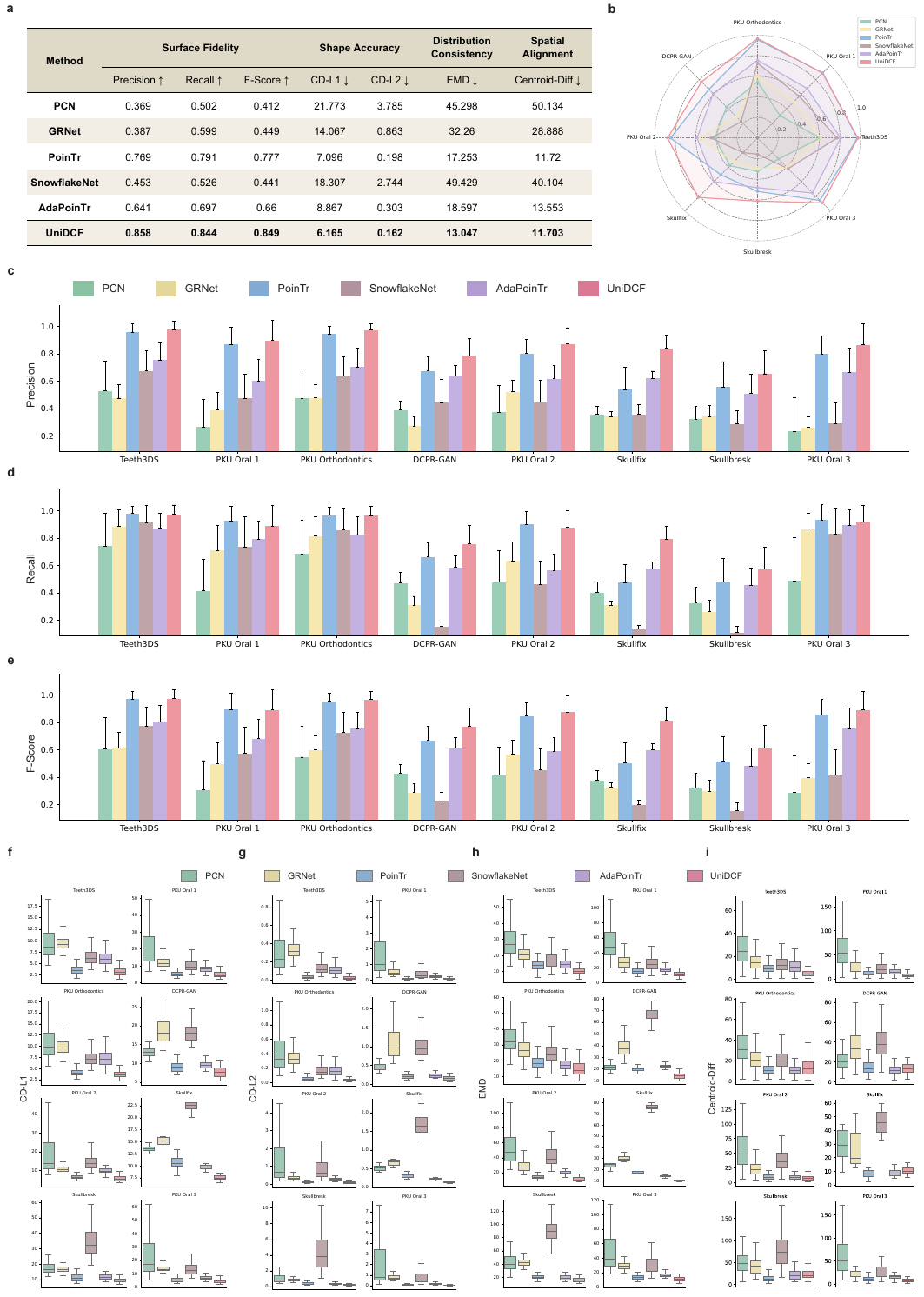}
\caption{
Comparative performance analysis between UniDCF and existing reconstruction methods.
a, Quantitative performance table comparing UniDCF with five baseline methods across eight datasets using all evaluation metrics.
b, Radar plot of per-dataset F1-Score comparing UniDCF and baseline methods.
c–e, Grouped bar charts illustrating performance in terms of Precision (c), Recall (d), and F1-Score (e) across all datasets.
f–i, Box plots showing distributions of four geometric metrics: CD-L1 (f), CD-L2 (g), EMD (h), and Centroid-Diff (i).}\label{fig4}
\end{figure}
To rigorously benchmark the performance and clinical applicability of UniDCF, we compared it with five current classic point cloud completion models: PCN \cite{PCN}, GRNet \cite{GRNet}, PoinTr \cite{PoinTr}, AdaPoinTr \cite{AdaPoinTr}, and SnowflakeNet \cite{SnowflakeNet}. These baselines span a spectrum of architectural paradigms, including classical point-based networks (PCN), voxel-aware contextual models (GRNet), and transformer-based frameworks (PoinTr, AdaPoinTr, and SnowflakeNet). All models were trained and evaluated under a unified experimental protocol using the integrated dentocraniofacial dataset to ensure fair comparison. We assessed model performance across four categories of quantitative metrics:  (1) Surface fidelity (Precision, Recall, and F1-score), measuring point-level accuracy and completeness within a fixed distance threshold (1\%); (2) Geometric accuracy (Chamfer Distance L1 [CD-L1] and L2 [CD-L2]), quantifying spatial discrepancy between predicted and reference point clouds; (3) Distributional consistency (Earth Mover’s Distance, EMD), evaluating global shape similarity; and (4) Spatial alignment (centroid difference, Centroid-Diff), measuring deviation between the centroids of predicted and ground truth geometries. For readability, all distance-based metrics (CD-L1, CD-L2, EMD, and Centroid-Diff) were multiplied by 1,000.

As shown in Fig.\ref {fig4}a, UniDCF consistently outperformed all baseline methods across all evaluation metrics, demonstrating superior geometric fidelity, structural completeness, and spatial accuracy. Compared with PCN and GRNet, UniDCF yielded substantial gains in all categories. Against transformer-based methods, UniDCF maintained a clear advantage. For instance, compared with the strongest transformer baseline—PoinTr—UniDCF achieved improvements of at least 0.53 in surface fidelity, 0.036 in shape accuracy, 4.206 in EMD, and 0.017 in centroid alignment error. Fig.\ref {fig4}b further illustrates UniDCF’s consistent and robust performance across datasets, in contrast to methods like PCN and SnowflakeNet, which showed greater sensitivity to dataset variability. As shown in Fig. \ref {fig4}c–e, UniDCF consistently outperformed all baseline methods in surface fidelity metrics, achieving higher precision, recall, and F1-scores across nearly all datasets, along with smaller standard deviations. In contrast with other models, such as SnowflakeNet—which exhibited sharp declines in recall on certain datasets—UniDCF demonstrated notably greater performance stability. This robustness is particularly evident in both recall and precision, which remained consistent across diverse anatomical contexts. Geometric distance metrics (Fig. \ref {fig4}f–h) further supported this trend: UniDCF yielded lower CD-L1, CD-L2, and EMD values than other methods, with median values concentrated near the lower bounds of their respective distributions. Centroid deviation analysis (Fig. \ref {fig4}i) also revealed more accurate and consistent spatial predictions across most datasets than other methods, underscoring UniDCF’s advantage in achieving precise anatomical alignment—an essential factor for clinical applications such as prosthetic fitting and surgical planning.

\begin{figure}[htbp]
\centering
\includegraphics[width=\textwidth]{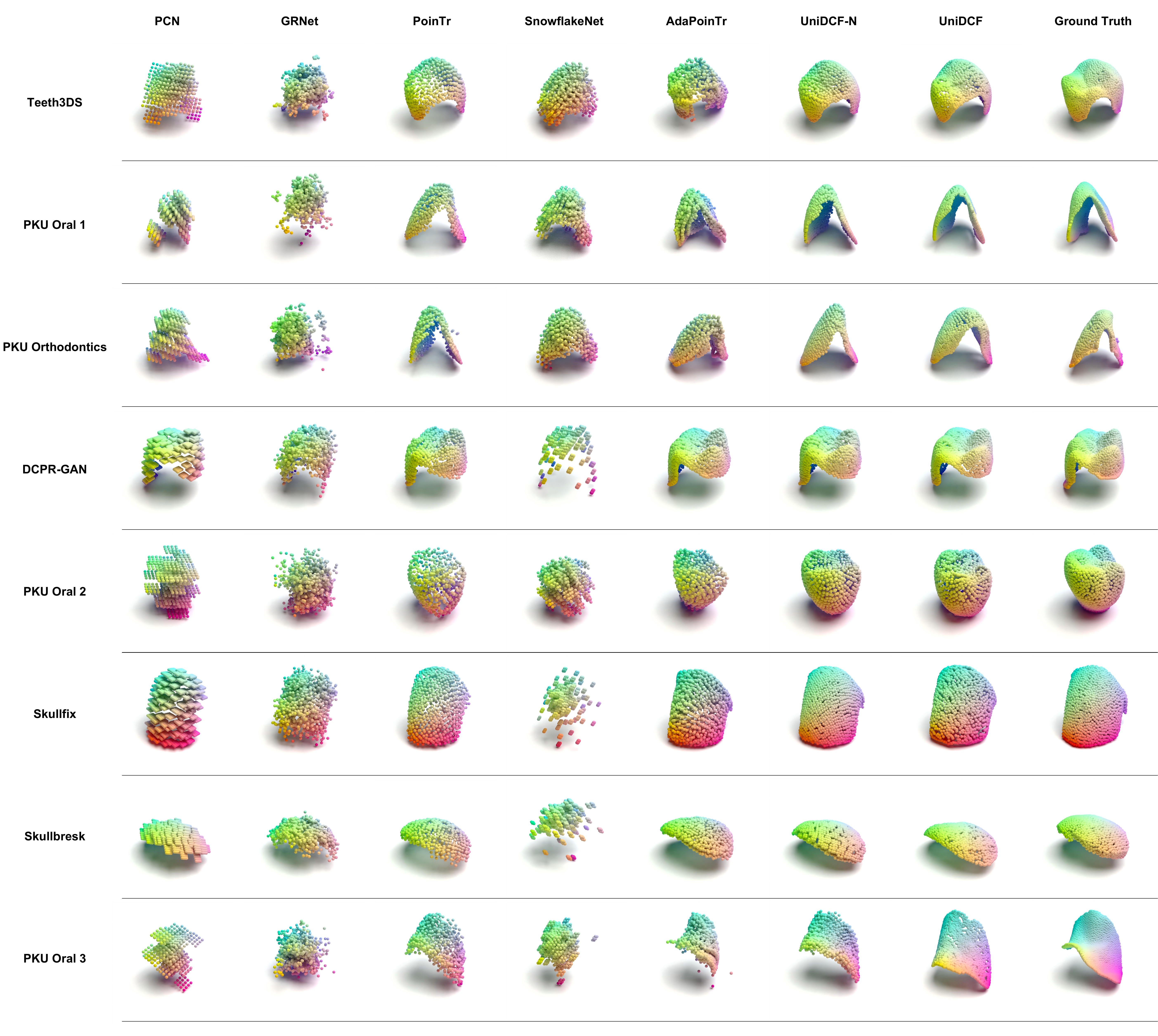}
\caption{Qualitative comparison of predicted point clouds across methods.
Columns 1–5 show reconstruction results from baseline methods. Column 6 displays results from our model using only multimodal fusion and point cloud completion (UniDCF-N). Column 7 shows full UniDCF results with score-based denoising. Column 8 provides the ground truth anatomical structures for visual comparison.}\label{fig5}
\end{figure}

Fig. \ref{fig5} provides qualitative comparisons across representative samples from all datasets. Traditional models (e.g., PCN and GRNet) often generated incomplete or noisy outputs, particularly in regions with complex morphology. SnowflakeNet preserved some dental crown details but performed poorly in craniofacial bone reconstructions, likely due to limited cross-domain generalizability. PoinTr and AdaPoinTr produced higher-quality reconstructions, especially in dental crown structures, reflecting their architectural similarity to UniDCF. UniDCF-N, a variant model by incorporating multimodal fusion and point completion but excluding denoising based on UniDCF, demonstrated improved geometric regularity and uniformity over baselines. Upon adding the score-based denoising module (full UniDCF), reconstructed geometries more closely approximated the ground truth. This resulted in visibly smoother surfaces and better-defined anatomical boundaries, particularly in challenging datasets such as Teeth3DS. Collectively, these results confirm that UniDCF surpasses existing methods in point cloud completion accuracy, especially in reconstructing fine-grained geometries and preserving structural detail in boundary regions.

\begin{figure}[htbp]
\centering
\includegraphics[width=\textwidth]{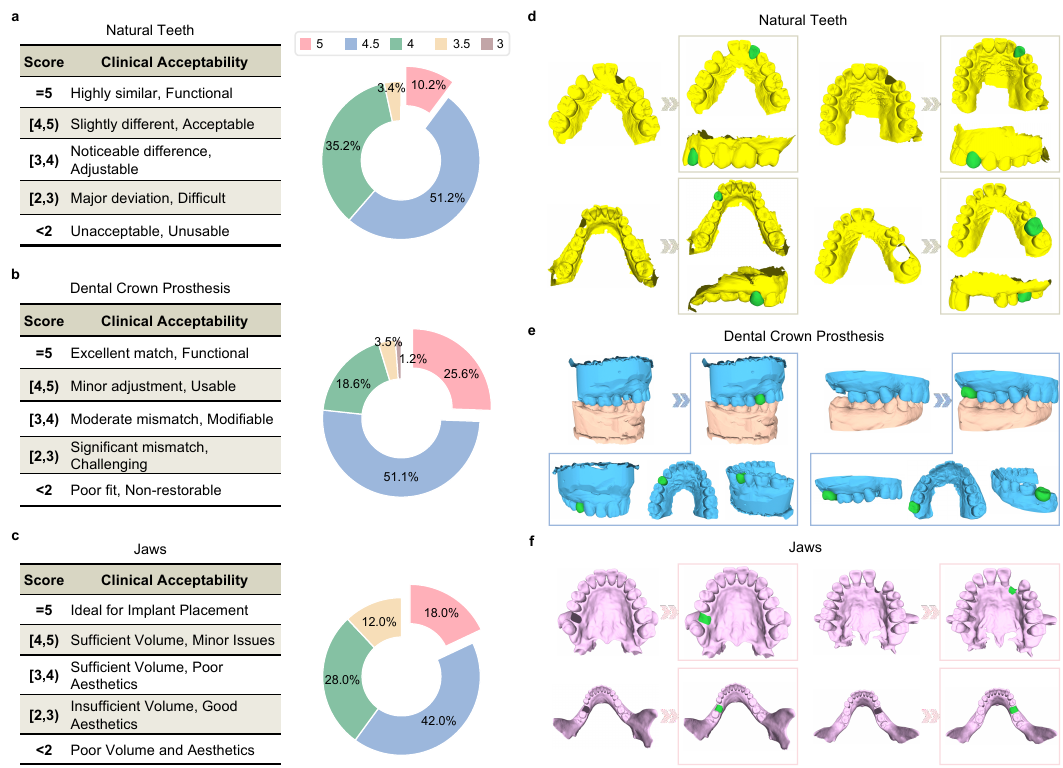}
\caption{Clinical evaluation and visualization of reconstructed results on clinical datasets. a–c, Clinical acceptability scores for reconstructions of natural teeth (a), dental crown prostheses (b), and jaws (c). Left panels show the evaluation rubric; right panels display score distributions as donut charts. A score of 5 indicates excellent acceptability; scores $\geqslant 4$ are considered clinically acceptable. d–f, Representative reconstruction examples for clinical evaluation:
d, Natural teeth reconstructions involving defects in the anterior teeth and posterior teeth regions, exhibiting varied arch morphologies and severity of crowding. e, Dental crown prostheses for both non-free-end and free-end edentulous spaces, accurately restored based on the geometry of adjacent dentition. f, Jaws reconstructions involving different anatomical sites in the maxilla and mandible.}\label{fig6}
\end{figure}

\subsubsection{Validation of Clinical Efficacy and Practical Value}\label{subsubsec223}
To evaluate the clinical applicability and effectiveness of UniDCF, we conducted a structured assessment on 224 randomly selected cases spanning three major dentocraniofacial hard tissue categories: natural teeth, dental crown prosthesis, and jaws. Three experienced clinicians independently rated the clinical acceptability of the model-generated reconstructions by comparing them against gold-standard anatomical references, including natural teeth, manually CAD designed dental crown prostheses, and conventional bone augmentation plans. Detailed scoring criteria are provided in Supplementary Appendix A.

UniDCF demonstrated a significant gain in design efficiency, generating restorative reconstructions in an average of 5.6 seconds per case. In contrast, manual design processes typically required 20–25 minutes for dental crowns and 15–45 minutes for craniofacial bone structures. This represents over 99\% reduction in planning time, highlighting the potential of UniDCF to streamline clinical workflows and alleviate clinician workload. As shown in Fig. \ref {fig6}a–c, the vast majority of reconstructions generated by UniDCF were rated as clinically acceptable or excellent. Across all three tissue types, an average of 94.2\% of cases received scores $\geqslant 4$, the predefined threshold for clinical acceptability. For dental crown prostheses, 95.3\% of cases surpassed the threshold, with 25.6\% receiving the highest score of 5. For natural teeth, 96.6\% of cases were rated $\geqslant 4$, including 10.2\% rated as perfect, underscoring the anatomical plausibility of the model's outputs. For jaws, 88.0\% of reconstructions were rated $\geqslant 4$, with 18.0\% achieving the maximum score. The high acceptability scores support its feasibility for direct clinical adoption or deployment with minimal manual refinement. Representative reconstruction cases used in the clinical evaluation are shown in Fig. \ref {fig6}d–f. For natural teeth (Fig. \ref {fig6}d), the model accurately reconstructed missing structures in the anterior and posterior regions of both dental arches. It generated anatomically plausible, personalized reconstructions across varying arch forms and severity of crowding. In dental crown prosthodontic (Fig. \ref {fig6}e), the model effectively restored crowns for both non-free-end and free-end edentulous cases, recovering anatomical morphology and occlusal function by leveraging geometric priors from adjacent teeth. For jaws tasks (Fig. \ref {fig6}f), UniDCF produced clinically accurate reconstructions across diverse anatomical subregions of the maxilla and mandible. Together with the clinician-rated evaluations, these examples further demonstrate UniDCF’s ability to generate anatomically faithful and clinically acceptable reconstructions across heterogeneous dentocraniofacial regions. 

\section{Discussion}\label{sec3}
\subsection{The Necessity of Large-Scale, Heterogeneous Datasets for Generalizable Multi-Tissue Reconstruction}\label{subsec31}
Our findings demonstrate that UniDCF, trained on a large-scale and anatomically diverse dataset, achieves significant performance gains in both teeth and craniofacial bone reconstruction tasks. Systematic comparisons revealed a clear performance hierarchy: the multi-task model consistently outperformed the single-task variants, which themselves surpassed models trained on isolated datasets. This progressive improvement underscores the critical role of dataset scale and task diversity in enhancing model robustness, generalizability, and clinical reliability. Despite the substantial anatomical differences between teeth and craniofacial bone structures, an interesting pattern emerged under multi-tissue training: performance gains were asymmetrical across tasks. Craniofacial bone reconstruction—limited by data availability—benefited markedly from the integration of teeth dataset, whereas teeth reconstruction—already supported by abundant task-specific dataset—maintained comparable performance regardless of the inclusion of craniofacial bone datasets. This asymmetry highlights the dynamics of cross-domain knowledge transfer: structural priors learned from large-scale teeth dataset can effectively support data-scarce tasks, while well-resourced tasks such as teeth reconstruction remain relatively unaffected by heterogeneous inputs. In addition, model performance also demonstrated sensitivity to dataset scale. Accuracy improved markedly when training dataset increased from 20\% to 60\%, with diminishing gains beyond 80\%, suggesting the saturation effect. This trend highlights the importance of strategic data curation and suggests that future improvements may rely more on complementary strategies—such as targeted data augmentation or enhanced representation learning—than on raw data volume alone. 

Importantly, cross-tissue dataset integration was particularly beneficial in data-scarce settings. Augmenting smaller craniofacial bone datasets with high-quality teeth dataset led to noticeable improvements, likely driven by the transfer of shared geometric priors and anatomical features. These findings support the clinical relevance of the proposed unified framework and its capacity to deliver robust, generalizable solutions across diverse real-world treatment scenarios.

\subsection{UniDCF for High-Fidelity Multimodal Reconstruction}\label{subsec32}
The effectiveness of UniDCF stems from its integration of two complementary modalities: sparse point clouds and multi-view grayscale images. Sparse point clouds provide accurate global geometry and preserve anatomical topology, but lack surface continuity. In contrast, multi-view grayscale images capture fine-grained surface detail and morphological consistency, yet offer limited 3D spatial accuracy. By fusing these modalities, UniDCF achieves both global structural coherence and localized geometric precision. Across all benchmark datasets, UniDCF consistently outperformed leading point cloud completion methods—including PCN, GRNet, PoinTr, AdaPoinTr, and SnowflakeNet across all evaluation metrics, validating the benefit of multimodal input fusion for anatomically complex reconstructions.

UniDCF is built upon the AdaPoinTr backbone—a geometry-aware, transformer-based encoder–decoder architecture optimized for point cloud completion. The transformer captures long-range dependencies to preserve global context, while the geometry-aware module encodes local spatial relationships. As shown in Fig. \ref{fig4}, transformer-based models (PoinTr, AdaPoinTr, UniDCF) outperformed non-transformer methods, particularly in regions with complex anatomical variation.
To further improve surface quality and clinical readiness, UniDCF incorporates a score-based denoising module. By treating predicted point clouds as samples from a noise-corrupted distribution, this module iteratively refines outputs via score-guided gradient ascent. As shown in Fig. \ref{fig5}, this approach significantly reduced surface irregularities and enhanced uniformity, yielding smoother reconstructions that better support downstream clinical workflows.
Taken together, the integration of multimodal fusion, transformer-based representation learning, and score-based denoising creates a high-performing pipeline. The modular design facilitates clinical transparency, enabling practitioners to understand and trust the reconstruction process, and highlights UniDCF’s potential for real-world adoption.

\subsection{Clinical Impact and Translational Outlook of a Unified Reconstruction Framework}\label{subsec33}
The clinical evaluation results confirm the strong translational potential of UniDCF. The model enables rapid dentocraniofacial reconstruction, reducing design time to approximately 5.6 seconds per case. Among 224 representative cases, 94.2\% of reconstructions were rated as clinically acceptable (score $\geqslant 4$) by experienced clinicians, with consistent performance across natural teeth, dental crown prosthesis, and jaws structures. These results highlight both the anatomical plausibility and robustness of UniDCF across heterogeneous tissue types. The visual quality and structural fidelity of the outputs further support the model’s readiness for routine clinical integration.

The dentocraniofacial hard tissues complex—including skull, teeth, and jaws regions—is characterized by significant anatomical heterogeneity, which presents a major challenge for precise, automated reconstruction. UniDCF addresses this challenge through a unified multimodal framework capable of accurately reconstructing structurally diverse hard tissues. Its generalizability across treatment scenarios—from prosthodontics to cranio-maxillofacial surgery and neurosurgery—underscores its potential for broad clinical application. In doing so, it reduces reliance on highly specialized expertise and enables more standardized, multidisciplinary care.

Nonetheless, several limitations warrant future attention. Clinically, further evaluation of the model on more complex, irregular, or large-scale defects is needed to assess its utility in advanced surgical planning. From a data perspective, future efforts should focus on collaborating with cranio-maxillofacial and neurosurgical specialists to construct a large-scale dataset of craniofacial bone defects, addressing the current imbalance wherein teeth data predominate despite the overall dataset size. Future work should explore direct point cloud-to-mesh reconstruction and incorporate explicit surface supervision to further enhance geometric fidelity and downstream clinical utility.

\section{Methods}\label{sec5}
\subsection{Data Preprocessing and Input Modalities}\label{subsec51}
\subsubsection{Data Preprocessing}\label{subsubsec511}
Dataset partitioning: Each dataset was randomly divided into training (90\%) and testing (10\%) subsets to ensure consistent evaluation across tasks.

Preprocessing of teeth datasets: 
In the Teeth3DS dataset, dental arches and individual teeth were automatically segmented using predefined positional labels, enabling the extraction of isolated dental crowns and partially defective arches. In the DCPR-GAN dataset, two trained researchers manually re-annotated each sample to explicitly define the relationships between edentulous regions and their corresponding missing dental crowns. For the PKU Oral 1 and PKU Orthodontics datasets, a pretrained intraoral scan segmentation model \cite{xiong2023tsegformer} was employed to process full-arch scans. Individual dental crowns were assigned distinct color-coded labels to ensure consistent identification and accurate segmentation.

Preprocessing of dental crown prosthesis datasets: Dental crown prosthesis datasets were retrospectively collected from patients who underwent implant treatment at the Peking University School and Hospital of Stomatology between August 2018 and August 2023. Inclusion criteria required a complete permanent dentition (first molar to second molar), excluding cases with mobile teeth or incomplete scans. Initial digital CAD-based dental crown design was created by intermediate-level prosthodontists following standardized protocols and subsequently reviewed and refined by two senior prosthodontists to ensure clinical validity.

Preprocessing of craniofacial bone datasets: A pretrained segmentation model \cite{isensee2021nnu} was applied to the Skullfix, Skullbreak, and PKU Oral 3 datasets to extract jaws and skull regions from CT or CBCT images. The resulting voxel data were converted into 3D meshes using the Marching Cubes algorithm \cite{mc}. Various severities of alveolar bone defects (mild, moderate, and severe) were manually simulated by an intermediate-level clinician and validated by two senior clinicians to ensure anatomical plausibility and clinical consistency.

Data governance and ethical compliance: The study received ethical approval from the Biomedical Ethics Committee of Peking University School and Hospital of Stomatology (approval no. PKUSSIRB-2024101112). Patient data were fully anonymized, excluding any identifiable personal information. Two dedicated researchers oversaw data management processes to ensure compliance with ethical standards and data security protocols.

\subsubsection{Construction of Input Modalities}\label{subsubsec512}
Following preprocessing, data from intraoral scans, CBCT, and CT were standardized into a unified coordinate system to eliminate inter-scan scale discrepancies. To enhance computational efficiency and reduce memory load, we applied farthest point sampling (FPS) to convert dense 3D meshes into sparse point clouds. This sampling approach preserved key geometric features while substantially reducing data dimensionality and redundancy. To capture localized geometric and topological features of defective regions, multi-view grayscale images were generated from the standardized 3D meshes. An automated rendering pipeline was developed using Blender, which converted normalized 3D models into 2D images from multiple predefined viewpoints. Renderings were generated from the +X, +Y, and +Z perspectives at a resolution of 256×256 pixels. To minimize modality redundancy and reduce computational overhead during training, RGB images were converted to grayscale while retaining essential spatial and structural information.

\subsection{Model Architecture and Algorithm Design}\label{subsec52}
We propose UniDCF, a unified framework for dentocraniofacial hard tissue reconstruction that integrates point cloud completion, multimodal feature fusion, and score-based denoising. 
Existing single-modality methods face fundamental limitations in capturing the complex geometry of dentocraniofacial structures. Pixel-based 2D approaches lack spatial depth and struggle to represent intricate 3D anatomy; voxel-based models are prohibitively memory-intensive for high-resolution reconstructions, particularly in craniofacia bones; and point-cloud-based methods, while computationally efficient, often fail to generalize across heterogeneous tissue types. In contrast, UniDCF leverages the complementary strengths of sparse point clouds—which provide accurate global spatial structure—and multi-view grayscale images—which encode rich local surface details—to produce anatomically faithful and clinically robust reconstructions.

\subsubsection{AdaPoinTr-based Point Cloud Completion Baseline}\label{subsubsec521}
The core point cloud completion component of our model is built upon AdaPoinTr \cite{AdaPoinTr}, selected for its geometry-aware Transformer-based encoder-decoder architecture. Transformer architectures inherently excel at modeling global contextual dependencies through self-attention mechanisms, allowing explicit encoding of structural knowledge and preserving fine-grained geometrical relationships. The geometry-aware module improves detailed reconstruction accuracy by explicitly modeling local geometric configurations. This rationale is quantitatively supported by visual and numerical comparisons where Transformer-based methods (PoinTr \cite{PoinTr}, AdaPoinTr, and ours) consistently reconstruct complex defect geometries with high accuracy (Fig. \ref{fig4} and Fig. \ref{fig5}).

Initially, the FPS is first applied to uniformly downsample the input into a point set containing a fixed number of points. We denote the partial input point cloud as 
$ \mathcal{P}_{in} = \{ p_i \mid i=1,2,...,N\}\in \mathbb{R}^{N\times3}$ where $N$ is the number of points and each point $p_i$ indicates a $(x, y, z)$ coordinate. 

A lightweight variant of the Dynamic Graph Convolutional Neural Network (DGCNN) \cite{dgcnn} is then employed as the point cloud feature extractor. For each point $p_i$, we obtain a point-wise feature vector $F_i^{'}$, which is further enhanced by a learnable point proxy representation defined as:
\begin{equation}
F_i = F_i^{'} + \phi(p_i),
\end{equation} 
where $\phi$ denotes a multilayer perceptron (MLP) applied to the 3D coordinates of $p_i$. The resulting set of point proxies $\mathcal{F} = \{ F_1, F_2, ..., F_N\}$ are subsequently fed into a geometry-aware Transformer encoder to produce a context-aware global structural representation $\mathcal{V}$. To accommodate defect regions with varying structural complexity, AdaPoinTr adopts an adaptive query generation module. The module generates query point embeddings $\mathcal{Q}$ based on the encoded global features 
$\mathcal{V}$, computed as follows:
\begin{equation}
\begin{aligned}
&\mathcal{Q} = MLP([\mathcal{C}, \mathcal{M}(Linear(\mathcal{V}))])\\
&\mathcal{C} = \mathcal{P}(Linear(\mathcal{V})),
\end{aligned}
\end{equation}
where $\mathcal{M}(\cdot)$ and $\mathcal{P}(\cdot)$ denote max pooling and coordinate projection operations, respectively, and $[\cdot, \cdot]$ represents feature concatenation. The query point embeddings $\mathcal{Q}$ and the encoder output $\mathcal{V}$ are jointly fed into a geometry-aware Transformer decoder, which progressively reconstructs the missing geometry through a series of multi-head attention layers. Finally, a high-resolution point cloud is generated using a reconstruction head based on FoldingNet \cite{FoldingNet}, which deforms a canonical 2D grid onto the underlying 3D surface to produce detailed geometric structures.

\subsubsection{Multimodal Encoding and Fusion}\label{subsubsec522}
The fusion module explicitly combines information from sparse point clouds and multi-view grayscale images to overcome the limitations of unimodal encoding. Sparse point clouds preserve global spatial relationships and coarse topology but lack surface continuity due to low sampling density. Conversely, grayscale images capture rich local curvature and contour features, though they lack direct 3D correspondence. By aligning and integrating these modalities, UniDCF achieves both global geometric accuracy and fine-grained anatomical realism—critical for clinically actionable reconstructions.

Specifically, the input point cloud $\mathcal{P}_{in}$ is processed by a geometry-aware Transformer encoder to extract geometric features $\mathcal{V}$. Concurrently, corresponding multi-view images $ \mathcal{X}_{in} = \{ x_i \mid i=1,2,...,N\}\in \mathbb{R}^{3 \times H \times W}$ are encoded via a pre-trained image feature extractor like DINO v2 model \cite{darcet2023vitneedreg, oquab2024dinov}, producing semantic image features $F_{img}$. Cross-modal fusion is achieved through a multi-head attention mechanism, spatially aligning features:
\begin{equation}
F_{fused} = \text{MultiHeadAttn}(\mathcal{V}, MLP(F_{img}), MLP(F_{img})) + \mathcal{V},
\end{equation}
where $MLP$ denotes a two-layer multilayer perceptron enhancing image feature embedding. The fused features $F_{fused}$ guide coarse-level reconstruction, which is subsequently refined through the geometry-aware Transformer decoder, resulting in improved anatomical accuracy.

\subsubsection{Score-based Point Cloud Denoising}\label{subsubsec523}
Although multimodal fusion improves overall reconstruction quality, reconstructed point clouds inevitably exhibit local noise and perturbations(manifested as irregularities or minor deviations from actual anatomical surfaces), negatively impacting clinical usability. To address this, we incorporated a score-based denoising method \cite{luo2021score}. This module treats the predicted point cloud as samples from a noise-perturbed distribution and estimates the gradient (score) of the log-probability density. By iteratively applying gradient ascent, the model refines the point cloud to recover smoother, anatomically coherent surfaces. Visual inspection (Fig. \ref{fig5}) confirms that this refinement reduces surface noise, yielding more uniform and clinically realistic reconstructions.

Formally, we modeled denoising as estimating the gradient of the log-probability density (score function):
\begin{equation}
\nabla_x \log[(p*n)(x)],
\end{equation}
where $p$ denotes the clean point cloud distribution representing true structural surfaces, and $n$ represents the noise distribution.
As depicted in Fig. \ref{fig1} c, a feature extractor based on densely connected dynamic graph convolutional layers extracts local and global geometric features from the noisy input points $X^{out}=\{x_i^{out}\}_{i=1}^N$. These extracted features, combined with local neighborhood coordinates, are input into a score estimation network to define local scores:
\begin{equation}
S_i(x^{out}) = \text{Score}(x^{out}-x_i^{out}, h_i),
\end{equation}
where $\text{Score}(\cdot)$ is MLP, $x^{out}$ represents neighborhood points, and $h_i$ denotes the geometric feature of $x_i^{out}$.
The network training objective aligns the predicted scores with ground-truth scores $s(x^{out})$, defined as vectors from noisy points to their nearest clean surfaces:
\begin{equation}
\mathcal{L}^{(i)} = \mathbb{E}_{x^{out}\sim\mathcal{N}(x_i^{out})}\left[\,\|s(x^{out}) - S_i(x^{out})\|_2^2\,\right],
\end{equation}
where $\mathcal{N}(x_i^{out})$ is a local neighborhood distribution in $\mathbb{R}^3$. The final training objective $\mathcal{L}$ aggregates each local score function objective $\mathcal{L}^{(i)}$:
\begin{equation}
\mathcal{L} = \frac{1}{N}\sum_{i=1}^N \mathcal{L}^{(i)}
\end{equation}
Finally, points were iteratively updated through gradient ascent:
\begin{equation}
\begin{aligned}
&{x_i^{out}}^{(t)} = {x_i^{out}}^{(t-1)} + \alpha_t \varepsilon_i({x_i^{out}}^{(t-1)}), \quad t=1,2,\dots,T \\
&{x_i^{out}}^{(0)} = {x_i^{out}}, {x_i^{out}}\in X^{out}
\end{aligned}
\end{equation}
where $\alpha_t$ is the step size at iteration $t$, and $\varepsilon_i$ is the ensemble score function aggregated from local neighborhoods. This denoising approach effectively eliminates irregularities without causing significant point shrinkage or distortion, resulting in smoother, more clinically relevant reconstructed surfaces.

\subsubsection{Evaluation Metrics}\label{subsubsec524}
To comprehensively evaluate the reconstruction performance of our model, we employed four groups of metrics designed to reflect critical aspects of geometric reconstruction quality.

(1) Surface coverage metrics (Precision, Recall, F-Score):
These metrics quantitatively assess the local accuracy and completeness of the predicted point clouds. Given a predicted point set $P$ and ground-truth point set $G$, the precision and recall are calculated at a predefined distance threshold $d$:
\begin{equation}
\begin{aligned}
\text{Precision} = \frac{\left|\{p \in \mathcal{P} \mid \min_{g \in \mathcal{G}}\|p - g\| < d\}\right|}{|\mathcal{P}|} \\
\text{Recall} = \frac{\left|\{g \in \mathcal{G} \mid \min_{p \in \mathcal{P}}\|g - p\| < d\}\right|}{|\mathcal{G}|}
\end{aligned}
\end{equation}
In our experiments, we set $d$ to 1\%. The F-Score combines precision and recall into a single balanced measure:
\begin{equation}
\text{F-Score} = 2 \times \frac{\text{Precision} \times \text{Recall}}{\text{Precision} + \text{Recall}}
\end{equation}

We selected these metrics to explicitly quantify how well predicted points cover true anatomical surfaces, crucial for clinical usability and reliability.

(2) Geometric accuracy metrics (Chamfer Distance L1 and L2):
Chamfer Distance measures the average nearest-neighbor distance between two point clouds, reflecting their geometric similarity. We employed two forms: the L1 norm (CD-L1) and the L2 norm (CD-L2):
\begin{equation}
\begin{aligned}
\text{CD-L1}(\mathcal{P}, \mathcal{G}) = \frac{1}{|\mathcal{P}|}\sum_{p\in \mathcal{P}}\min_{g\in \mathcal{G}}\|p - g\|_1 + \frac{1}{|\mathcal{G}|}\sum_{g\in \mathcal{G}}\min_{p\in \mathcal{P}}\|g - p\|_1 \\
\text{CD-L2}(\mathcal{P}, \mathcal{G}) = \frac{1}{|\mathcal{P}|}\sum_{p\in \mathcal{P}}\min_{g\in \mathcal{G}}\|p - g\|_2^2 + \frac{1}{|\mathcal{G}|}\sum_{g\in \mathcal{G}}\min_{p\in \mathcal{P}}\|g - p\|_2^2
\end{aligned}
\end{equation}
CD-L1 emphasizes overall geometric fidelity, while CD-L2 penalizes larger deviations more severely, ensuring robust evaluation of structural precision and highlighting significant reconstruction errors relevant to clinical decision-making.

(3) Point distribution consistency metric (Earth Mover’s Distance, EMD):
The EMD quantifies the minimum total displacement required to align the predicted point cloud distribution with the ground truth distribution, thus evaluating global spatial consistency:
\begin{equation}
\text{EMD}(\mathcal{P}, \mathcal{G}) = \min_{\phi:\mathcal{P} \rightarrow \mathcal{G}}\frac{1}{|\mathcal{P}|}\sum_{p\in \mathcal{P}}\|p - \phi(p)\|
\end{equation}
where $\phi$ denotes a bijective mapping between point clouds. EMD was chosen to ensure uniformity and global spatial coherence of predicted reconstructions, vital for accurate anatomical restoration and clinical usability.

(4) Centroid difference (Spatial localization accuracy):
To explicitly quantify spatial alignment, we computed centroid differences as the Euclidean distance between predicted ($C_P$) and ground-truth ($C_G$) point cloud centroids:
\begin{equation}
\text{Centroid-Diff} = \|C_\mathcal{P} - C_\mathcal{G}\|_2,\quad\text{where}\quad C_X=\frac{1}{|X|}\sum_{x\in X}x
\end{equation}
This metric directly evaluates the model’s capability to correctly localize reconstructed anatomy, an essential factor in clinical precision and surgical applicability.

Collectively, these metrics were carefully selected to provide comprehensive and clinically meaningful evaluations of our reconstruction method, reflecting accuracy, structural fidelity, spatial consistency, and practical utility.

\subsubsection{Computational Environment}\label{subsubsec525}
Experiments were conducted on an Ubuntu server equipped with single NVIDIA A100 GPU, implemented using Python 3.9 and the PyTorch deep learning framework.

\subsection{Model validation and Statistical analysis}\label{subsec53}
\subsubsection{Evaluating Cross-task Generalization Driven by Multi-tissue Dataset Diversity}\label{subsec531}
To systematically assess the impact of heterogeneous dataset diversity on cross-task generalization, we reformulated dental crown restoration and craniofacial bone reconstruction as two tasks within the broader domain of hard tissue reconstruction. We compared three experience settings: (1) independent training on individual datasets (Single-dataset), (2) joint training on task-specific datasets grouped by tissue type (teeth-only or craniofacial bone-only), and (3) unified multi-tissue training using the entire multimodel dataset (All dataset). Validation was conducted on standardized testing sets, with CD-L1 and F1-Score serving as the primary evaluation metrics. To further assess the influence of dataset scale, we trained models using incremental subsets (20\% to 100\%) of the training data to identify performance saturation thresholds.

\subsubsection{Comparative and Visual Analysis}\label{subsec532}
To benchmark UniDCF against existing approaches, we compared it with several point cloud completion methods, including PCN \cite{PCN}, GRNet \cite{GRNet}, PoinTr, AdaPoinTr, and SnowflakeNet \cite{SnowflakeNet}. Evaluation was performed comprehensively across four dimensions: surface fidelity (Precision, Recall, F1-Score), shape accuracy (CD-L1, CD-L2), distribution consistency (EMD), and spatial alignment (Centroid-Diff). All methods were evaluated on the same datasets to ensure fairness. Visual comparisons of reconstructed point clouds were also conducted across all datasets to qualitatively assess geometric plausibility and anatomical consistency.

\subsubsection{Validation of Clinical Efficacy and Practical Value}\label{subsec534}
To assess the clinical feasibility of UniDCF, we conducted a validation study on 224 randomly selected cases encompassing natural teeth, dental crown prostheses, and jaws structures. Three experienced clinicians independently rated the acceptability of the reconstructions using predefined clinical criteria. In parallel, efficiency was measured by comparing the average time required for model-generated versus manually designed restorations. The results demonstrate that UniDCF’s potential for significantly enhancing clinical workflow efficiency and practicality in routine teeth and craniofacial bone reconstruction.

\backmatter
\bmhead{Acknowledgements}
This research was supported by Beijing Natural Science Foundation (L232113, L242134, L242059, L252199, L252162) and the Fundamental Research Funds for the Central Universities(2023RC09).

\bmhead{Author contribution}
C.-X.R., H.S., Y.Z. and L.X. conceived the idea. L.X.,Y.Z. and H.S. designed and supervised the experiments. C.-X.R., N.Z. and Y.L. implemented and performed the experiments. Y.-Y.H., S.-Y.L. and S.-W.M. analysed the data and experimental results. N.Z., Y.L., G.C., R.-J.W., H.S. and Y.Z. annotated the data and conducted clinical trial analysis. C.-X.R., N.Z.,L.X. wrote the paper. All authors commented on the paper.

\begin{appendices}

\section{Detailed Clinical Rating Criteria for Subjective Acceptability Evaluation}\label{secA1}
\begin{figure}[htbp]
\centering
\includegraphics[width=\textwidth]{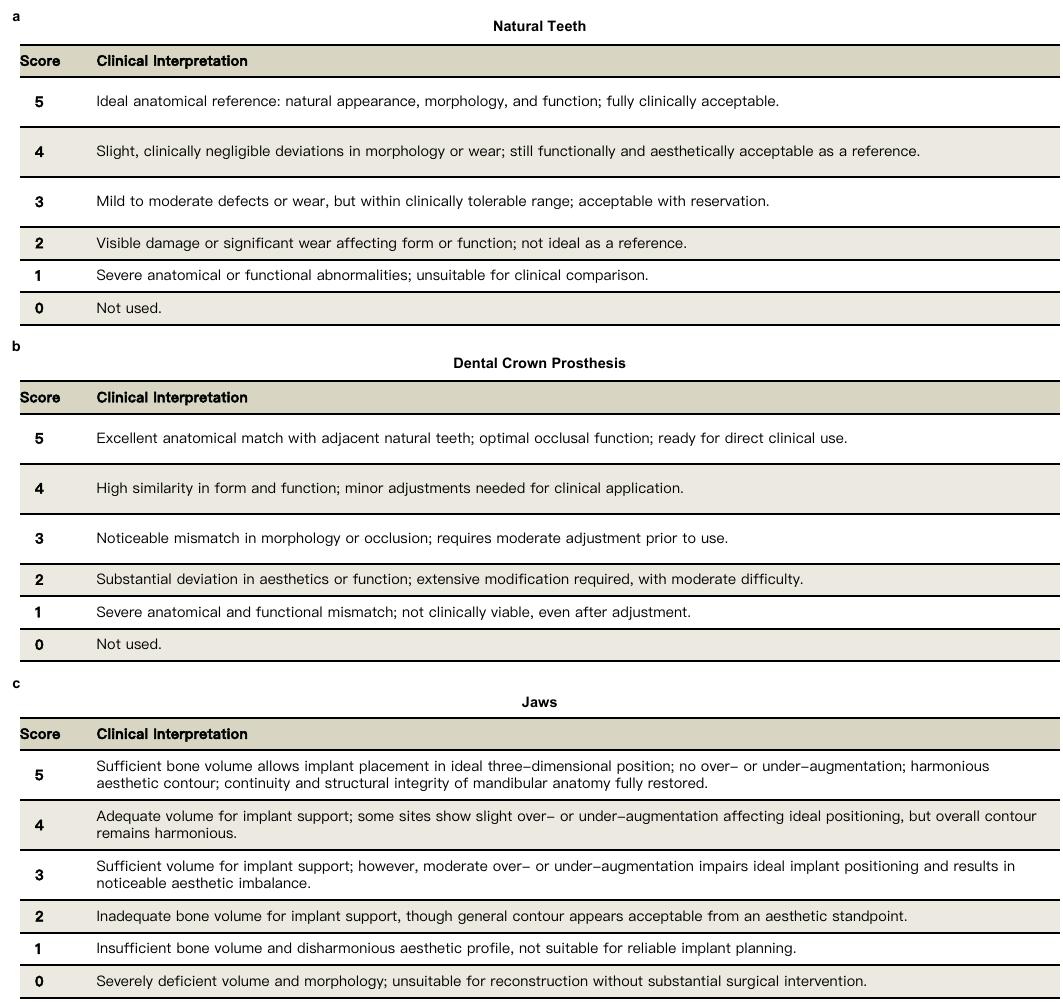}
\caption{The detailed clinical criteria to assess the acceptability of model-generated restorations, and consistent with Fig. \ref{fig6}a-c.}\label{fulu1}
\end{figure}

To complement the summary scores presented in Fig. \ref{fig6}a-c, this appendix (Fig. \ref{fulu1}) provides the detailed clinical criteria used by five experienced evaluators to assess the acceptability of model-generated restorations. Each restoration was rated on a six-point scale (0–5), and scores were grouped into five clinical relevance intervals consistent with Fig. \ref{fig6}a-c:

(1) [0–2): Poor fit / Unacceptable

(2) [2–3): Significant deviation / Challenging

(3) [3–4): Moderate mismatch / Modifiable

(4) [4–5): Minor adjustment / Acceptable

(5) =5: Ideal match / Fully functional




\end{appendices}

\bibliography{sn-bibliography.bib}

\end{document}